\pdfoutput=1

\documentclass[11pt]{article}

\usepackage[]{acl}

\usepackage{times}
\usepackage{latexsym}

\usepackage{amsmath}
\usepackage{amsfonts}
\usepackage{gensymb}
\usepackage{subcaption}
\usepackage{booktabs}
\newcommand{\ra}[1]{\renewcommand{\arraystretch}{#1}}
\newcommand{\act}[1]{{\tt\MakeUppercase{#1}}}
\usepackage{graphicx} 

\usepackage[T1]{fontenc}

\usepackage[utf8]{inputenc}

\usepackage{microtype}

\title{Analyzing Generalization of Vision and Language Navigation to Unseen Outdoor Areas}

\author{Raphael Schumann \\
  Computational Linguistics\\
  Heidelberg University, Germany \\
  \And
  Stefan Riezler \\
  Computational Linguistics \& IWR \\
  Heidelberg University, Germany \\
  \hspace{-7cm}\texttt{\{rschuman|riezler\}@cl.uni-heidelberg.de}}

\begin{document}
\maketitle
\begin{abstract}
Vision and language navigation (VLN) is a challenging visually-grounded language understanding task. Given a natural language navigation instruction, a visual agent interacts with a graph-based environment equipped with panorama images and tries to follow the described route. 
Most prior work has been conducted in indoor scenarios where best results were obtained for navigation on routes that are similar to the training routes, with sharp drops in performance when testing on unseen environments. We focus on VLN in outdoor scenarios and find that in contrast to indoor VLN, most of the gain in outdoor VLN on unseen data is due to features like junction type embedding or heading delta that are specific to the respective environment graph, while image information plays a very minor role in generalizing VLN to unseen outdoor areas. These findings show a bias to specifics of graph representations of urban environments, demanding that VLN tasks grow in scale and diversity of geographical environments.\footnote{Code: \url{https://github.com/raphael-sch/map2seq_vln} \\ Data \& Demo: \url{https://map2seq.schumann.pub/vln/}}
\end{abstract}

\section{Introduction}
Vision and language navigation (VLN) is a challenging task that requires the agent to process natural language instructions and ground them in a visual environment. The agent is embodied in the environment and receives navigation instructions. Based on the instructions, the observed surroundings, and the current trajectory the agent decides its next action. Executing this action changes the position and/or heading of the agent within the environment, and eventually the agent follows the described route and stops at the desired goal location. The most common evaluation metric in VLN is the proportion of successful agent navigations, called task completion (TC). 

While early work on grounded navigation was confined to grid-world scenarios~\cite{walkthetalk, interpret_nlni}, recent work has studied VLN in outdoor environment consisting of real-world urban street layouts and corresponding panorama pictures~\cite{Chen2018Touchdown}. Recent agent models for outdoor VLN treat the task as a sequence-to-sequence problem where the instructions text is the input and the output is a sequence of actions~\cite{Chen2018Touchdown, Xiang2020LearningNavigation, Zhu2020MultimodalNavigation}. In contrast to indoor VLN~\cite{room2room, rxr}, these works only consider a \textit{seen scenario}, i.e., the agent is tested on routes that are located in the same area as the training routes. However, studies of indoor VLN~\cite{Zhang2020DiagnosingNavigation} show a significant performance drop when testing in previously unseen areas. 

The main goal of our work is to study \textit{outdoor VLN in unseen areas}, pursuing the research question of which representations of an environment and of instructions an agent needs to succeed at this task. We compare existing approaches to a new approach that utilizes features based on the observed environment graph to improve generalization to unseen areas. The first feature, called junction type embedding, encodes the number of outgoing edges at the current agent position; the second feature, called heading delta, encodes the agent's heading change relative to the previous timestep. As our experimental studies show, representations of full images do not contribute very much to successful VLN in outdoor scenarios beyond these two features. One reason why restricted features encoding junction type and heading delta are successful in this task is that they seem to be sufficient to encode peculiarities of the graph representation of the environments. Another reason is the current restriction of outdoor environments to small urban areas. In our case, one dataset is the widely used Touchdown dataset introduced by \citet{Chen2018Touchdown}, the other dataset is called map2seq and has recently been introduced by \citet{schumann-riezler-2021-map2seq}. The map2seq dataset was created for the task of navigation instructions generation but can directly be adopted to VLN. We conduct a detailed analysis of the influence of general neural architectures, specific features such as junction type or heading delta, the role of image information and instruction token types, to outdoor VLN in seen and unseen environments on these two datasets.

Our specific findings unravel the contributions of these features on several VLN subtasks such as orientation, directions, stopping. Our general finding is that current outdoor VLN suffers a bias towards urban environments and to artifacts of their graph representation, showing the necessity of more diverse datasets and tasks for outdoor VLN.

Our main contributions are the following:
 \begin{itemize}
     \item We describe a straightforward agent model that achieves state-of-the-art task completion and is used as a basis for our experiments.
     \item We introduce the \textit{unseen scenario} for outdoor VLN and propose two environment-dependent features to improve generalization in that setting.
     \item We compare different visual representations and conduct language masking experiments to study the effect in the unseen scenario.
     \item We adopt the map2seq dataset to VLN and show that merging it with Touchdown improves performance on the respective test sets.
\end{itemize}

\section{VLN Problem Definition}
\label{sec:problem}
The goal of the agent is to follow a route and stop at the desired target location based on natural language navigation instructions. The environment is a directed graph with nodes~$v~\in~\mathbb{V}$ and labeled edges~$(u,v)~\in~\mathbb{E}$. Each node is associated with a 360\degree~panorama image~$p$ and each edge is labeled with an angle~$\alpha_{(u,v)}$. The agent state~$s~\in~S$ consists of a node and the angle at which the agent is heading:~$(v, \alpha_{(v,u)}~|~u\in~\mathbb{N}_v^{out})$, where $\mathbb{N}_v^{out}$ are all outgoing neighbors of node $v$. The agent can navigate the environment by performing an action $a \in \{\act{forward}, \act{left}, \act{right}, \act{stop}\}$ at each timestep $t$. The $\act{forward}$ action moves the agent from state~$(v, \alpha_{(v,u)})$ to~$(u, \alpha_{(u,u')})$, where~$(u,u')$ is the edge with an angle closest to $\alpha_{(v,u)}$. The $\act{right}$ and $\act{left}$ action rotates the agent towards the closest edge angle in clockwise or counterclockwise direction, respectively:~$(v, \alpha_{(v,u')})$. Given a starting state~$s_1$ and instructions text~$\mathbf{x}$, the agent performs a series of actions $a_1,...,a_T$ until the $\act{stop}$ action is predicted. If the agent stops within one neighboring node of the desired target node (goal location), the navigation was successful. The described environment and location finding task was first introduced by \cite{Chen2018Touchdown} and we will also refer to it as "outdoor VLN task" throughout this paper. 

\section{Model Architecture}
\label{model}
\begin{figure}[t]
    \centering
    \includegraphics[width=0.49\textwidth]{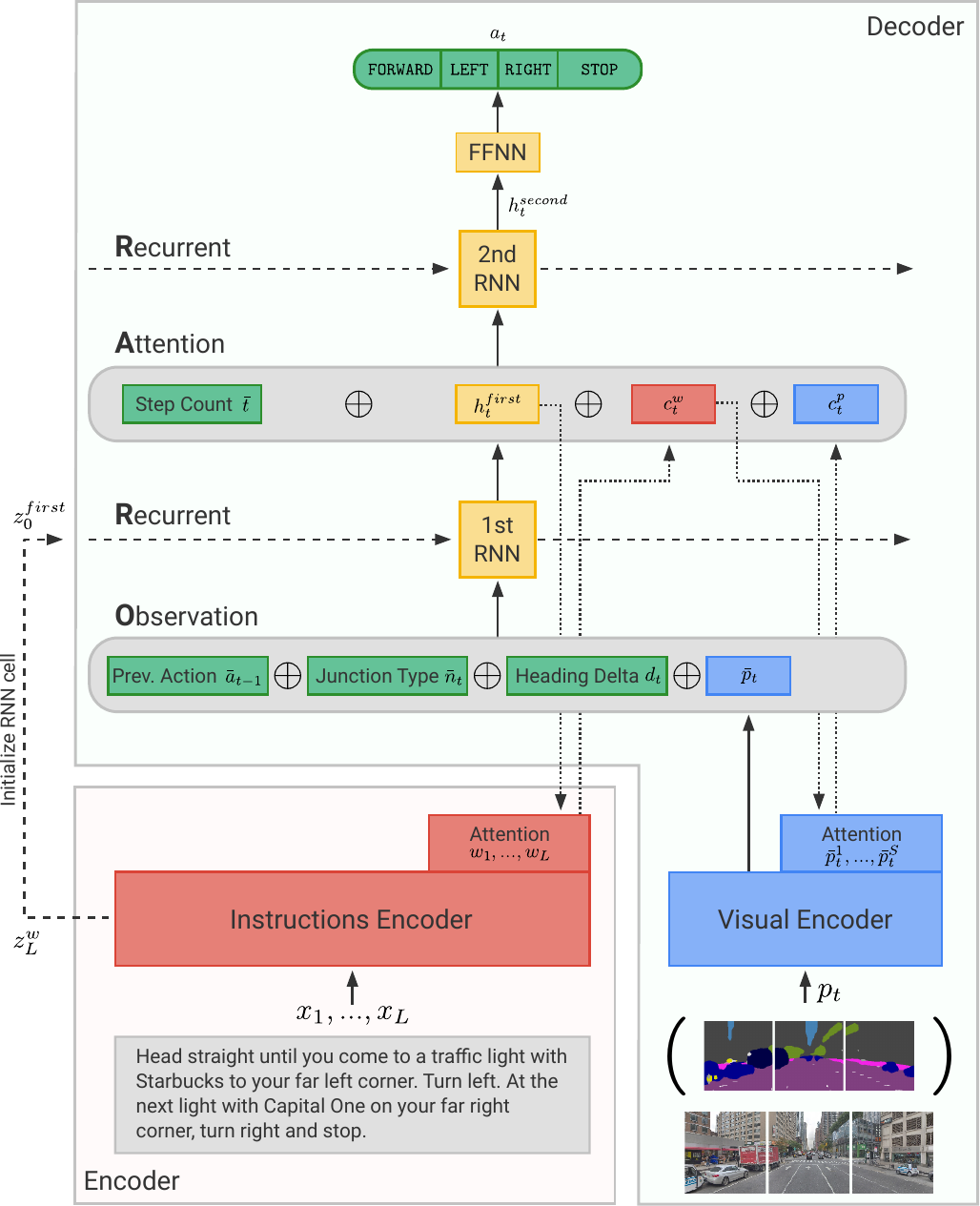}
    \caption{The ORAR model for outdoor vision and language navigation follows a sequence-to-sequence architecture. The instructions text is encoded and used along the visual features to predict the next agent action. The recurrent decoder has two layers, the first encodes observations about the current environment state, the second allows attention over the input text and panorama view. The predicted action changes the state of the agent in the environment and with it the panorama view of the next timestep.}
    \label{fig:model}
\end{figure}

In this section we introduce the model that we use to analyze navigation performance in the unseen and seen scenario for outdoor VLN. The architecture is inspired by the cross-modal attention model for indoor VLN~\citep{vln_ce}. First we give a high level overview of the model architecture and rough intuition. Afterwards we provide a more formal description.

As depicted in Figure \ref{fig:model}, the model follows a sequence-to-sequence architecture where the input sequence is the navigation instructions text and the output is a sequence of agent actions. At each decoding timestep, a new visual representation of the current agent state within the environment is computed, where the agent state is dependent on the previously predicted actions.
The decoder RNN has two layers where the first encodes metadata and a visual representation. The second RNN layer encodes a contextualized text and visual representation and eventually predicts the next action.

The intuition behind the model architecture is to firstly accumulate plain \textit{observations} available at the current timestep and entangle them with previous observations in the first \textit{recurrent} layer. Based on these observations, the model focuses \textit{attention} to certain parts of the instructions text and visual features which are again entangled in the second \textit{recurrent} layer. Thus, we use the acronym \textit{ORAR} (observation-recurrence attention-recurrence) for the model.

In detail, the instructions encoder embeds and encodes the tokens in the navigation instructions sequence $\mathbf{x}=x_1, ..., x_L$ using a bidirectional LSTM \cite{bilstm}:
\begin{align*}
\hat{x}_i&=\text{embedding}(x_i)\\
((w_1,...,w_L), z^w_L)&=\text{Bi-LSTM}(\hat{x}_1,...,\hat{x}_L),
\end{align*}
where $w_1,...,w_L$ are the hidden representations for each token and $z^w_L$ is the last LSTM cell state. The visual encoder, described in detail below, emits a fixed size representation $\bar{p}_t$ of the current panorama view and a sequence of sliced view representations $\bar{p}_t^1,...,\bar{p}_t^S$. The state $z^{first}_0$ of the cell in the first decoder LSTM layer is initialized using $z^w_L$. The input to the first decoder layer is the concatenation ($\oplus$) of visual representation $\bar{p}_t$, previous action embedding $\bar{a}_{t-1}$, junction type embedding $\bar{n}_t$, and heading delta $d_t$. The output of the first decoder layer,
\begin{equation*}
h^{first}_t=\text{LSTM}^{first}([\bar{a}_{t-1} \oplus \bar{n}_t \oplus d_t \oplus \bar{p}_t]),
\end{equation*}
is then used as the query of multi-head attention~\cite{vaswani-etal-2017-attention} over the text encoder. The resulting contextualized text representation $c^w_t$ is then used to attend over the sliced visual representations:
\begin{align*}
c^w_t&=\text{MultiHeadAttention}(h^{first}_t,(w_1,...,w_L))\\
c^p_t&=\text{MultiHeadAttention}(c^w_t,(\bar{p}_t^1,...,\bar{p}_t^S)).
\end{align*}
The input and output of the second decoder layer are
\begin{equation*}
h^{second}_t=\text{LSTM}^{second}([\bar{t} \oplus h^{first}_t \oplus c^w_t \oplus c^p_t]),
\end{equation*}
where $\bar{t}$ is the embedded timestep $t$. The hidden representation $h^{second}_t$ of the second decoder LSTM layer is then passed through a feed forward network to predict the next agent action $a_t$.
\subsection{Visual Encoder}
At each timestep $t$ the panorama at the current agent position is represented by
extracted visual features. We slice the panorama into eight projected rectangles with 60\degree~field of view, such that one of the slices aligns with the agent's heading. This centering slice and the two left and right of it are fed into a ResNet pretrained\footnote{\url{https://pytorch.org/vision/0.8/models.html}} on ImageNet~\cite{imagenet}. We consider two variants of ResNet derived panorama features. One variant extracts low level features from the fourth to last layer~(\textbf{4th-to-last}) of a pretrained ResNet-18 and concatenates each slice's feature map along the width dimension, averages the 128 CNN filters and cuts out 100 dimensions around the agents heading. This results in a feature matrix of $100 \times 100$~($\bar{p}_t^1,...,\bar{p}_t^{100}$). The full procedure is described in detail in \citet{Chen2018Touchdown} and \citet{Zhu2020MultimodalNavigation}. The other variant extracts high level features from a pretrained \mbox{ResNet-50's} \textbf{pre-final} layer for each of the 5 slices:~$\bar{p}_t^1,...,\bar{p}_t^{5}$. Each slice vector $\bar{p}_t^s$ is of size $2,048$ resulting in roughly the same number of extracted ResNet features for both variants, making a fair comparison. Further, we use the \textbf{semantic~segmentation} representation of the panorama images. We employ omnidirectional semantic segmentation~\cite{yang2020omnisupervised} to classify each pixel by one of the 25 classes of the Mapillary Vistas dataset~\cite{mapillary}. The classes include e.g. car, truck, traffic light, vegetation, road, sidewalk. See Figure \ref{fig:model} bottom right for a visualization. Each panorama slice~($\bar{p}_t^1,...,\bar{p}_t^{5}$) is then represented by a 25 dimensional vector where each value is the normalized area covered by the corresponding class~\cite{Zhang2020DiagnosingNavigation}. For either feature extraction method, the fixed sized panorama representation~$\bar{p}_t$ is computed by concatenating the slice features~$\bar{p}_t^1,...,\bar{p}_t^{S}$ and passing them to a feed forward network.

\subsection{Junction Type Embedding}
\label{sec:junction}
The junction type embedding is a feature that we introduce to better analyze generalization to unseen areas. It embeds the number of outgoing edges of the current environment node and is categorized into~\{2,~3,~4,~\textgreater4\}. It provides the agent information about the type of junction it is positioned on: a regular street segment, a three-way intersection, a four way intersection or an intersection with more than four outgoing streets. We want to point out that the number of outgoing edges isn't oracle information in the environment described in Section \ref{sec:problem}. The agent can rotate left until the same panorama view is observed and thus counting the number of outgoing edges by purely interacting with the environment. But it is clear that the feature leverages the fact that the environment is based on a graph and it would not be available in a continuous setting~\cite{vln_ce}.

\subsection{Heading Delta}
\label{sec:heading}
\begin{figure}
\centering
  \includegraphics[width=0.35\textwidth]{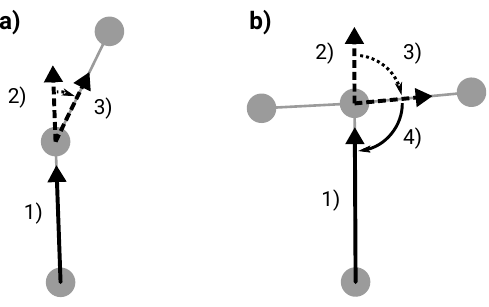}
  \caption{Visualization of automatic agent rotation initiated by the environment. Grey circles and interconnecting edges are part of the environment graph. Black solid arrows are actions initiated by the agent. Black dotted arrows depict agent heading and automatic rotation by the environment.
  \textbf{a)}: 1) The agent moves forward.~2)~Agent's heading does not point to an outgoing edge. 3) Agent is automatically rotated to the closest edge without causing problems.
  \textbf{b)}: The agent receives instructions like "Turn right at the next intersection".~1)~The agent moves forward. 2) Agent's heading does not point to an outgoing edge. 3) The environment automatically rotates the agent towards the closest outgoing edge. 4)~The agent has no explicit information about the automatic rotation and predicts a right turn as instructed, leading to a failed navigation.}
  \label{fig:heading}
\end{figure}
As described in Section \ref{sec:problem}, the environment defined and implemented by \citet{Chen2018Touchdown} only allows states where the agent is heading towards an outgoing edge. As a consequence the environment automatically rotates the agent towards the closest outgoing edge after transitioning to a new node. The environment behavior is depicted in Figure \ref{fig:heading}a) for a transition between two regular street segments. 
However, as depicted in Figure \ref{fig:heading}b), a problem arises when the agent is walking towards a three-way intersection. The automatic rotation introduces unpredictable behavior for the agent and we hypothesis that it hinders generalization to unseen areas. To correct for this environment artifact, we introduce the heading delta feature $d_t$ which encodes the change in heading direction relative to the previous timestep. The feature is normalized to~$(-1,1]$ where a negative value indicates a left rotation and a positive value indicates a right rotation. The magnitude signals the degree of the rotation up to 180\degree.

\section{Data}
\label{sec:data}
We use the Touchdown~\cite{Chen2018Touchdown} and the map2seq~\cite{schumann-riezler-2021-map2seq} datasets in our experiments. Both datasets contain human written navigation instructions for routes located in the same environment. The environment consists of 29,641 panorama images from Manhattan and the corresponding connectivity graph.

\subsection{Touchdown}
The Touchdown dataset~\cite{Chen2018Touchdown} for vision and language navigation consists of 9,326 routes paired with human written navigation instructions. The annotators navigated the panorama environment based on a predefined route and wrote down navigation instructions along the way. 
\subsection{Map2seq}
The map2seq~\cite{schumann-riezler-2021-map2seq} dataset was created for the task of navigation instructions generation. The 7,672 navigation instructions were written by human annotators who saw a route on a rendered map, without the corresponding panorama images. The annotators were told to include visual landmarks like stores, parks, churches, and other amenities into their instructions. A different annotator later validated the written navigation instructions by using them to follow the described route in the panorama environment~(without the map). This annotation procedure allows us to use the navigation instructions in the map2seq dataset for the vision and language navigation task. We are the first to report VLN results on this dataset.

\subsection{Comparison}
Despite being located in the same environment, the routes and instructions from each dataset differ in multiple aspects. The map2seq instructions typically include named entities like store names, while Touchdown instructions focus more on visual features like the color of a store. Both do not include street names or cardinal directions and are written in egocentric perspective.  Further, in map2seq the agent starts by facing in the correct direction, while in Touchdown the initial heading is random and the first part of the instruction is about orientating the agent ("Turn around such that the scaffolding is on your right"). A route in map2seq includes a minimum of three intersections and is the shortest path from the start to the end location.\footnote{The shortest path bias reduces the number of reasonable directions at each intersection and thus makes the task easier.} In Touchdown there are no such constraints and a route can almost be circular. The routes in both datasets are around 35-45 nodes long with some shorter outliers in Touchdown. On average instructions are around 55 tokens long in map2seq and around 89 tokens long in Touchdown.

\section{Experiments}
We are interested in the generalization ability to unseen areas and how it is influenced by the two proposed features, types of visual representation, navigation instructions and training set size. Alongside of the results in the unseen scenario, we report results in the seen scenario to interpret performance improvements in relation to each other. All experiments\footnote{Except comparison models on the Touchdown seen test set for which we copy the results from the respective work.} are repeated ten times with different random seeds. The reported numbers are the average over the ten repetitions. Results printed in \textbf{bold} are significantly better than non-bold results in the same column. Significance was established by a paired t-test\footnote{\url{https://docs.scipy.org/doc/scipy/reference/generated/scipy.stats.ttest_rel.html}} on the ten repetition results and a p-value ~$\leq0.05$ without multiple hypothesis corrections factor. Individual results can be found in the Appendix. 
\subsection{Data Splits}
\label{data_splits}
\begin{figure}
    \centering
    \begin{subfigure}{0.21\textwidth}
        \includegraphics[width=\textwidth]{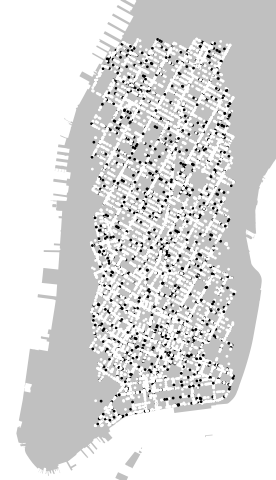}
    \end{subfigure}
    \hfill
    \begin{subfigure}{0.21\textwidth}
        \includegraphics[width=\textwidth]{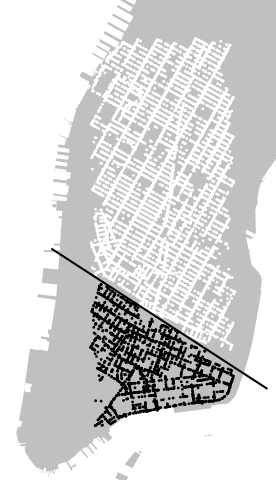}
    \end{subfigure}
    \caption{Visualization of the environment area located in Manhattan. \textbf{The seen scenario is depicted on the left and the unseen scenario on the right.} Each white dot is a training route and each black dot is a test route in the Touchdown and map2seq dataset. The unseen scenario is characterized by geographic separation of the training and testing area.}
  \label{fig:splits}
\end{figure}
To be able to compare our model with previous work, we use the original training, development and test split \cite{Chen2018Touchdown} for the seen scenario on Touchdown. Because we are the first to use the map2seq data for VLN we create a new split for it. The resulting number of instances can be seen in the left column of Table \ref{tab:data}. For the unseen scenario, we create new splits for both datasets. We separate the unseen area geographically by drawing a boundary across lower Manhattan (see Figure \ref{fig:splits}). Development and test instances are randomly chosen from within the unseen area. Routes that are crossing the boundary are discarded. The right column of Table~\ref{tab:data} shows the number of instances for both splits. Additionally, we merge the two datasets for both scenarios. This is possible because both datasets are located in the same environment and the unseen boundary is equivalent.
\begin{table}[]
\centering
\resizebox{.99\linewidth}{!}{
\begin{tabular}{@{}lrrrcrrr@{}} 
\toprule
          & \multicolumn{3}{c}{\textbf{seen}}  & \phantom{} & \multicolumn{3}{c}{\textbf{unseen}}            \\ 
\cmidrule{2-4} \cmidrule{6-8}
          & \textbf{train} & \textbf{dev} & \textbf{test} && \textbf{train} & \textbf{dev} & \textbf{test}  \\ 
\midrule
Touchdown & 6,525           & 1,391         & 1,409          && 6,770           & 800          & 1,507           \\
map2seq   & 6,072           & 800          & 800           && 5,737           & 800          & 800            \\
\midrule
Merged   & 12,597           & 2,191          & 2,209           && 12,507           & 1,600          & 2,307            \\
\bottomrule
\end{tabular}
}
\caption{Number of instances in the data splits for the seen and unseen scenario of Touchdown and map2seq.}
\label{tab:data}
\end{table}

\begin{table*}[]
\centering
\resizebox{.99\linewidth}{!}{
\begin{tabular}{@{}lc@{}cc@{}clc@{}cc@{}clc@{}cc@{}clc@{}cc@{}c@{}}
\toprule
    &   \multicolumn{9}{c}{\textbf{Seen}} & \phantom{} &\multicolumn{9}{c}{\textbf{Unseen}}            \\ 
\cmidrule{2-10} \cmidrule{12-20}
    &   \multicolumn{4}{c}{\textbf{Touchdown}} & \phantom{} & \multicolumn{4}{c}{\textbf{map2seq}} && 
        \multicolumn{4}{c}{\textbf{Touchdown}} & \phantom{} & \multicolumn{4}{c}{\textbf{map2seq}} \\ 
\cmidrule{2-5} \cmidrule{7-10} \cmidrule{12-15} \cmidrule{17-20}
    &   \multicolumn{2}{c}{\textbf{dev}} & \multicolumn{2}{c}{\textbf{test}} && \multicolumn{2}{c}{\textbf{dev}} &         \multicolumn{2}{c}{\textbf{test}} && \multicolumn{2}{c}{\textbf{dev}} & \multicolumn{2}{c}{\textbf{test}} &&        \multicolumn{2}{c}{\textbf{dev}} & \multicolumn{2}{c}{\textbf{test}} \\ 
    \midrule
    \textbf{Model} & \multicolumn{1}{r}{\small{nDTW}} & TC & \multicolumn{1}{c}{\small{nDTW}} & TC && \multicolumn{1}{c}{\small{nDTW}} & TC & \multicolumn{1}{c}{\small{nDTW}} & TC && \multicolumn{1}{c}{\small{nDTW}} & TC & \multicolumn{1}{c}{\small{nDTW}} & TC && \multicolumn{1}{c}{\small{nDTW}} & TC & \multicolumn{1}{c}{\small{nDTW}} & TC \\
\midrule
RConcat         & 22.5 & 10.6 & 22.9 & 11.8 && 30.7 & 17.1 & 27.7 & 14.7 && 3.9 & 2.3 & 3.5 & 1.9 && 3.7 & 2.0 & 3.8 & 2.1 \\
GA              & 25.2 & 12.0 & 24.9 & 11.9 && 33.0 & 18.2 & 30.1 & 17.0 && 3.6 & 1.8 & 4.0 & 2.2 && 3.9 & 1.8 & 4.1 & 1.7 \\
ARC             & - & 15.3 & - & 14.1 && - & - & - & - && - & - & - & - && - & - & - & - \\
ARC+l2s         & - & 19.5 & - & 16.7 && - & - & - & - && - & - & - & - && - & - & - & - \\
VLN Transformer & 23.0 & 14.0 & 25.3 & 14.9 && 31.1 & 18.6 & 29.5 & 17.0 && 4.7 & 2.3 & 5.2 & 3.1 && 6.2 & 3.6 & 6.1 & 3.5 \\
\midrule
ORAR full model &&&&&&&&&&&&&&&&&&&\\
$\bullet$ ResNet pre-final & 38.9 & 26.0 & 38.4 & 25.3 && \textbf{65.0} & 49.1 & \textbf{62.3} & \textbf{46.7} && 13.0 & 9.6 & 12.1 & 8.8 && 34.6 & 24.2 & 34.5 & 24.6 \\
$\bullet$ ResNet 4th-to-last & \textbf{45.1} & \textbf{29.9} & \textbf{44.9} & \textbf{29.1} && 60.0 & 43.4 & 57.8 & 41.7 && \textbf{22.2} & \textbf{15.4} & \textbf{21.6} & \textbf{14.9} && \textbf{41.0} & \textbf{27.6} & \textbf{42.2} & \textbf{30.3} \\
\midrule
ORAR full model&\multicolumn{4}{c}{$\bullet$ ResNet 4th-to-last} & \phantom{} & \multicolumn{4}{c}{$\bullet$ ResNet pre-final} & \phantom{} & \multicolumn{4}{c}{$\bullet$ ResNet 4th-to-last} & \phantom{} & \multicolumn{4}{c}{$\bullet$ ResNet 4th-to-last} \\
- no heading delta & \textbf{45.5} & \textbf{30.0} & \textbf{45.3} & \textbf{29.3} && 63.2 & 47.7 & \textbf{60.3} & 44.9 && \textbf{21.6} & \textbf{15.2} & \textbf{21.2} & \textbf{14.8} && 33.0 & 22.0 & 33.6 & 23.6 \\
- no junction type & 40.6 & 25.9 & 40.9 & 25.5 && \textbf{65.9} & \textbf{52.9} & \textbf{62.1} & \textbf{47.5} && 7.9 & 4.8 & 7.1 & 4.3 && 13.1 & 7.4 & 11.8 & 7.1 \\
- no head. \& no junc. & 39.2 & 24.6 & 39.4 & 24.2 && 62.7 & \textbf{49.6} & 58.9 & 45.1 && 7.6 & 4.6 & 7.0 & 4.4 && 8.9 & 5.0 & 8.2 & 4.7 \\
\bottomrule
\end{tabular}
}
\caption{Results on Touchdown and map2seq for the seen and unseen scenario. Metrics are normalized Dynamic Time Warping~(nDTW) and task completion~(TC). In the first section we list results for the comparison models: RConcat, GA, VLN Transformer~\cite{Zhu2020MultimodalNavigation} and ARC, ARC+learn2stop~\cite{Xiang2020LearningNavigation}. In the second section we present results for the ORAR model with two different types of image features: \textit{ResNet pre-final} features are extracted from the last layer before the classification and \textit{ResNet 4th-to-last} are low level features extracted from the fourth to last layer of a pretrained ResNet. The last section ablates the two proposed features: \textit{heading delta} and \textit{junction type embedding}.}
\label{tab:results}
\end{table*}

\subsection{Training Details}
We train the models with Adam~\cite{adam} by minimizing cross entropy loss in the teacher forcing paradigm. We set the learning rate to 5e-4, weight decay to 1e-3 and batch size to 64. After 150 epochs we select the model with the best shortest path distance (SPD) performance on the development set. We apply dropout of 0.3 after each dense layer and recurrent connection. The multi-head attention mechanism is regularized by attention dropout of 0.3 and layer normalization. The navigation instructions are lower-cased and split into byte pair encodings~\cite{bpe} with a vocabulary of 2,000 tokens and we use BPE dropout~\cite{bpe_dropout} during training. The BPE embeddings are of size 32 and the bidirectional encoder LSTM has two layers of size 256. The feed forward network in the visual encoder consists of two dense layers with 512 and 256 neurons, respectively, and 64 neurons in case of using semantic segmentation features. The embeddings that encode previous action, junction type, and step count are of size 16. The two decoder LSTM layers are of size 256 and we use two attention heads. Training the full model takes around 3 hours on a GTX~1080~Ti.

\subsection{Model Comparison}
We compare the ORAR model to previous works. Because these works only report results for the seen scenario on Touchdown, we evaluate those for which we could acquire the code, on the map2seq dataset and the unseen scenario. The models \textit{RConcat}~\cite{city_without_map, Chen2018Touchdown}, \textit{GA}~\cite{gated_attention, Chen2018Touchdown} and \textit{ARC}~\cite{Xiang2020LearningNavigation} use an LSTM to encode the instructions text and a single layer decoder LSTM to predict the next action. They differ in how the text and image representations are incorporated during each timestep in the decoder. As the name suggests, in \textit{RConcat} the two representations are concatenated. \textit{GA} uses gated attention to compute a fused representation of text and image. \textit{ARC} uses the hidden representation of the previous timestep to attend over the instructions text. This contextualized text representation is then concatenated to the image representation. They further introduce \textit{ARC+l2s} which cascades the action prediction into a binary stopping decision and a subsequent direction classification. The \textit{VLN-Transformer} \cite{Zhu2020MultimodalNavigation} uses pretrained BERT \cite{devlin-etal-2019-bert} to encode the instructions and VLN-BERT \cite{web_image_pairs} to fuse the modalities.

\subsection{Metrics}
We use task completion~(\textbf{TC}) as the main performance metric. It represents the percentage of successful agent navigations~\cite{Chen2018Touchdown}. We further report normalized~Dynamic Time Warping~(\textbf{nDTW}) which quantifies agent and gold trajectory overlap for all routes~\cite{ndtw}. The shortest path distance~(\textbf{SPD}) is measured within the environment graph from the node the agent stopped to the goal node~\cite{Chen2018Touchdown}.

\section{Results \& Analysis}
The two upper sections of Table~\ref{tab:results} show the results of the ORAR model introduced in Section~\ref{model} in comparison to other work. While the model significantly outperforms all previous work on both datasets, our main focus is analyzing generalization to the unseen scenario. It is apparent that the type of image features influences agent performance and will be discussed in the next section. The bottom section of Table~\ref{tab:results} ablates the proposed heading delta and junction type features for the best models. Removing the heading delta feature has little impact in the seen scenario, but significantly reduces task completion in the unseen scenario of the map2seq dataset. Surprisingly, the feature has no impact in the unseen scenario of Touchdown. We believe this is a consequence of the different data collection processes. Touchdown was specifically collected for VLN and annotators navigated the environment graph, while map2seq annotators wrote instructions only seeing the map. Removing the junction type embedding leads to a collapse of task completion in the unseen scenario on both datasets. This shows that without this explicit feature, the agent lacks the ability to reliably identify intersections in new areas. 

\subsection{Visual Features}
\begin{table}[]
\centering
\resizebox{.9\linewidth}{!}{
\begin{tabular}{lcclcc}
\toprule
& \multicolumn{5}{c}{\textbf{Unseen}}\\
\cmidrule{2-6}
& \multicolumn{2}{c}{\textbf{Touchdown}} & \phantom{} & \multicolumn{2}{c}{\textbf{map2seq}} \\
\cmidrule{2-3}\cmidrule{5-6}
\textbf{Visual Features} & \textbf{dev} & \textbf{test} && \textbf{dev} & \textbf{test} \\
\midrule
ResNet pre-final        &  9.6 &  8.8 && 24.2 & 24.6 \\
- no junction type      &  4.4 &  4.0 && 10.7 & 11.0 \\
\midrule
ResNet 4th-to-last      & \textbf{15.4} & \textbf{14.9} && 27.6 & \textbf{30.3} \\
- no junction type      &  \textbf{4.8} &  4.3 &&  7.4 & 7.1 \\
\midrule
semantic segmentation   & 11.5 & 11.0 && \textbf{29.0} & \textbf{31.1} \\
- no junction type      & \textbf{5.5} & \textbf{5.5} && \textbf{11.6} & \textbf{12.1} \\
\midrule
no image                & 11.5 & 9.5 && \textbf{28.5} & \textbf{30.5} \\
- no junction type      & 3.0 & 2.8 && 5.4 & 5.5 \\
\bottomrule
\end{tabular}
}
\caption{Study of visual features for the unseen scenario of Touchdown and map2seq. Metric is task completion.}
\label{tab:visual_features}
\end{table}

Table \ref{tab:visual_features} shows results for different types of visual features in the unseen scenario. We compare high level ResNet features (pre-final), low level ResNet features (4th-to-last), semantic segmentation features and using no image features. For the ResNet based features, the low level 4th-to-last features perform better than pre-final on both datasets. On map2seq the no image baseline performs on par with models that have access to visual features. When we remove the junction type embedding, the task completion rate drops significantly, which shows that the agent is not able to reliably locate intersections from any type of visual features.

\subsection{Sub-task Oracle}
\label{sec:oracle}
\begin{table}[]
\centering
\resizebox{.8\linewidth}{!}{
\begin{tabular}{lcclcc}
\toprule
& \multicolumn{5}{c}{\textbf{Touchdown}}\\
\cmidrule{2-6}
     & \multicolumn{2}{c}{\textbf{Seen}} & \phantom{} & \multicolumn{2}{c}{\textbf{Unseen}} \\
\cmidrule{2-3}\cmidrule{5-6}
    \textbf{Sub-task} & \textbf{dev}            & \textbf{test}          && \textbf{dev}          & \textbf{test}         \\
\midrule
ORAR pre-final           & 26.0 & 25.3 && 9.6 & 8.8 \\
\midrule
orientation             & 79.2 & 77.5 && 66.7 & 67.6 \\
directions           & \underline{84.8} & \underline{85.5} && 45.9 & 45.7 \\
stopping             & \underline{40.7} & \underline{41.0} && \underline{37.4} & \underline{36.1} \\
\midrule
ORAR 4th layer           & \textbf{29.9} & \textbf{29.1} && \textbf{15.4} & \textbf{14.9} \\
\midrule
orientation             & \underline{92.4} & \underline{91.5} && \underline{84.2} & \underline{84.1} \\
directions           & 81.6 & 81.1 && 53.4 & 52.4 \\
stopping             & \underline{39.7} & \underline{40.2} && \underline{36.4} & \underline{35.2} \\
\midrule
ORAR no image        & 15.2 & 13.3 && 11.1 & 9.5 \\
\midrule
orientation             & 59.8 & 57.0 && 61.3 & 60.5 \\
directions           & 74.1 & 73.3 && \underline{58.8} & \underline{57.9} \\
stopping             & \underline{39.3} & \underline{38.8} && \underline{36.1} & 34.0\\
\bottomrule
\end{tabular}
}
\caption{Oracle analysis on Touchdown. Division into three sub-tasks: \textit{orientation}, \textit{directions} and \textit{stopping}. Providing oracle actions for two of the three sub-tasks allows an isolated look at the remaining one. Underlined results are best for the sub-task, e.g. 85.5 is the best TC for the directions task on the test set in the seen scenario.}
\label{tab:oracle} 
\end{table}

The agent has to predict a sequence of actions in order to successfully reach the goal location. In Touchdown this task can be divided into three sub-tasks (see Section~\ref{sec:data}). First the agent needs to \textbf{orientate} itself towards the correct starting heading. Next the agent has to predict the correct \textbf{directions} at the intersections along the path. The third sub-task is \textbf{stopping} at the specified location. Providing oracle actions (during testing) for two of the three sub-tasks lets us look at the completion rate of the remaining sub-task. Table~\ref{tab:oracle} shows the completion rates for each of the three sub-tasks when using ResNet pre-final, 4th-to-last and no image features. In the seen scenario we can observe that the pre-final features lead to the best performance for the directions task. The 4th-to-last features on the other hand lead to the best orientation task performance and the stopping task is not influenced by the choice of visual features. In the unseen scenario 4th-to-last features again provide best orientation task performance but no image features lead to the best performance for the directions task. This shows that the ResNet~4th-to-last features are primarily useful for the orientation sub-task and explains the discrepancy of the no image baseline on Touchdown and map2seq identified in the previous subsection. In the Appendix we use this knowledge to train a mixed-model that uses 4th-to-last features for the orientation sub-task and pre-final/no image features for directions and stopping.

\subsection{Token Masking}
\begin{figure}
\centering
  \includegraphics[width=0.49\textwidth]{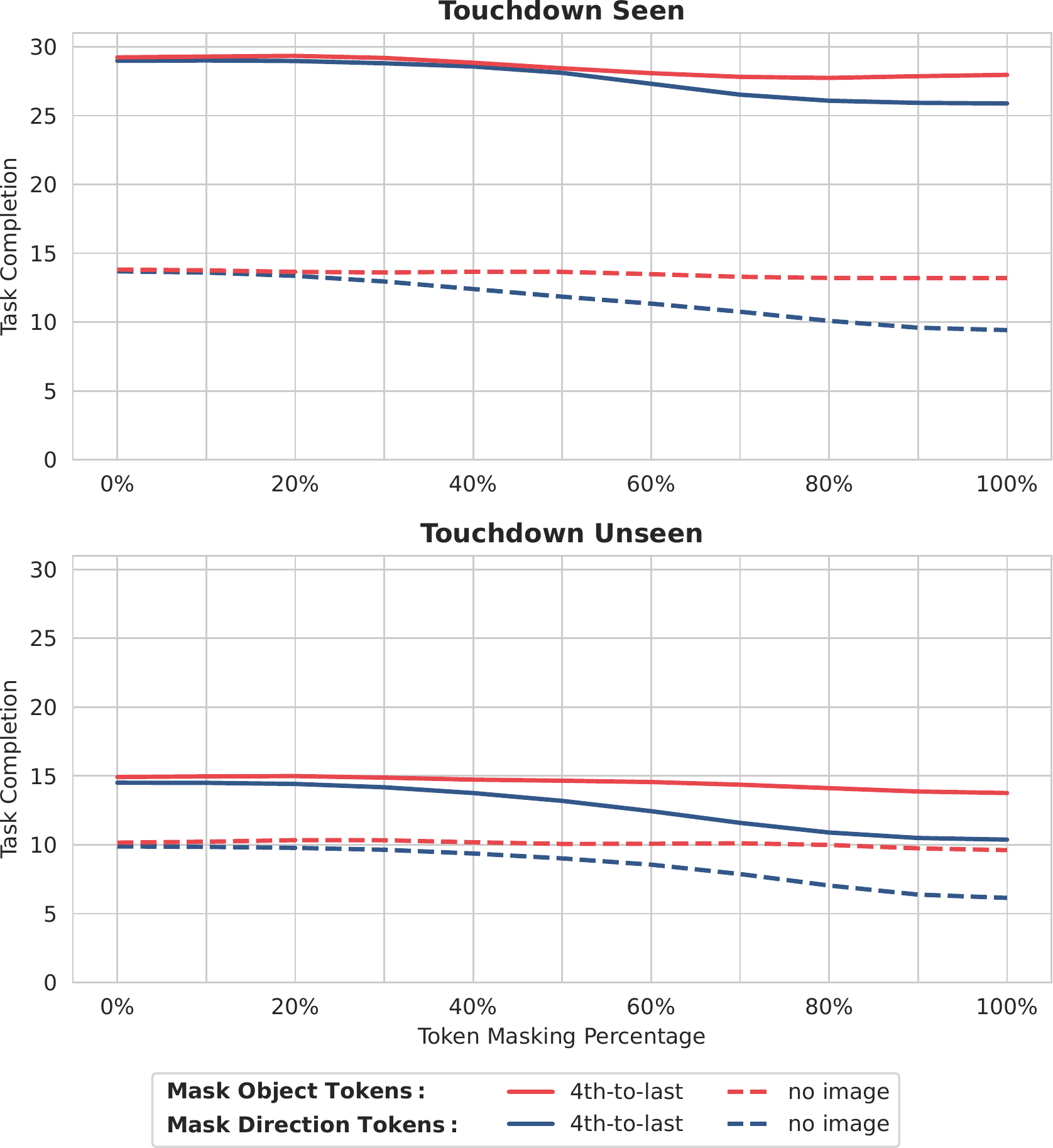}
  \caption{Masking experiments on the seen and unseen test set of Touchdown. Object or direction tokens are masked during training and testing.}
  \label{fig:masking}
\end{figure}
To analyze the importance of direction and object tokens in the navigation instructions, we run masking experiments similar to~\citet{what_really_matters}, except that we mask the tokens during training and testing instead of during testing only. Figure~\ref{fig:masking} shows the resulting task completion rates for an increasing number of masked direction or object tokens. From the widening gap between masking object and direction tokens, we can see that the direction tokens are more important to successfully reach the goal location. Task completion nearly doesn't change when masking object tokens, indicating that they are mostly ignored by the model. While task completion significantly drops when direction tokens are masked, the agent still performs on a high level. This finding is surprising and in dissent with~\citet{what_really_matters} who report that task completion nearly drops to zero when masking direction tokens during testing only. We believe that in our setting (masking during testing and training), the model learns to infer the correct directions from redundancies in the instructions or context around the direction tokens. Besides the general trend of lower performance on the unseen scenario, we can not identify different utilization of object or direction tokens in the seen and unseen scenario.

\subsection{Merged Datasets}
\begin{table*}[ht]
\centering
\resizebox{.99\linewidth}{!}{
\begin{tabular}{@{}lc@{}cc@{}clc@{}cc@{}clc@{}cc@{}clc@{}cc@{}c@{}}
\toprule
    &   \multicolumn{9}{c}{\textbf{Seen}} & \phantom{} &\multicolumn{9}{c}{\textbf{Unseen}}            \\ 
    
\cmidrule{2-10} \cmidrule{12-20}

    & \multicolumn{4}{c}{\textbf{Merged}} & \phantom{} & \multicolumn{2}{c}{\textbf{Touchdown}} & \multicolumn{2}{c}{\textbf{map2seq}} && 
       \multicolumn{4}{c}{\textbf{Merged}} & \phantom{} & \multicolumn{2}{c}{\textbf{Touchdown}} & \multicolumn{2}{c}{\textbf{map2seq}} \\ 
       
\cmidrule{2-5} \cmidrule{7-10} \cmidrule{12-15} \cmidrule{17-20}

    &  \multicolumn{2}{c}{\textbf{dev}} & \multicolumn{2}{c}{\textbf{test}} && \multicolumn{2}{c}{\textbf{test}} & \multicolumn{2}{c}{\textbf{test}} &&         \multicolumn{2}{c}{\textbf{dev}} & \multicolumn{2}{c}{\textbf{test}} && \multicolumn{2}{c}{\textbf{test}} & \multicolumn{2}{c}{\textbf{test}} \\ 
    
    \midrule
    
    \textbf{Model} & \multicolumn{1}{r}{\small{nDTW}} & TC & \multicolumn{1}{c}{\small{nDTW}} & TC && \multicolumn{1}{c}{\small{nDTW}} & TC & \multicolumn{1}{c}{\small{nDTW}} & TC && \multicolumn{1}{c}{\small{nDTW}} & TC & \multicolumn{1}{c}{\small{nDTW}} & TC && \multicolumn{1}{c}{\small{nDTW}} & TC & \multicolumn{1}{c}{\small{nDTW}} & TC \\
\midrule
best non-merged & - & - & - & - && \textbf{44.9} & \textbf{29.1} & 62.3 & 46.7 && - & - & - & - && 21.6 & 14.9 & 42.2 & 30.3 \\

\midrule
ORAR full model\\
$\bullet$ no image          & 37.5 & 26.6 & 35.8 & 24.7 && 23.0 & 14.8 & 58.3 & 42.1 && 31.6 & 22.3 & 27.0 & 19.2 && 16.6 & 11.7 & \textbf{46.5} & \textbf{33.2} \\
$\bullet$ ResNet pre-final  & 51.3 & \textbf{38.8} & 49.3 & \textbf{36.8} && 39.1 & 27.7 & \textbf{67.3} & \textbf{52.8} && 28.9 & 22.0 & 25.7 & 20.0 && 17.4 & 13.6 & 41.3 & 32.1 \\
$\bullet$ ResNet 4th-to-last  & \textbf{53.4} & \textbf{37.8} & \textbf{51.8} & \textbf{35.7} && \textbf{46.0} & \textbf{30.1} & 62.1 & 45.5 && \textbf{35.7} & \textbf{25.4} & \textbf{33.6} & \textbf{24.2} && \textbf{27.0} & \textbf{19.3} & \textbf{46.1} & \textbf{33.5} \\
\bottomrule
\end{tabular}
}
\caption{Results for models trained on the merged dataset. Test results are presented for the merged test set and individual Touchdown and map2seq test sets. Metrics are normalized Dynamic Time Warping~(nDTW) and task completion~(TC). In the first row the best results of Table \ref{tab:results} (non-merged training sets) are listed for comparison. The bottom section presents results on the \textit{ORAR full model} with different types of image features.}
\label{tab:merged}
\end{table*}

We train the ORAR full model on the merged dataset (see Section~\ref{data_splits}). Model selection is performed on the merged development set but results are also reported for the individual test sets of Touchdown and map2seq. For comparison with models trained on the non-merged datasets, the first row of Table \ref{tab:merged} shows the best results of Table \ref{tab:results}. Training on the merged dataset significantly improves nDTW and task completion across both datasets and scenarios. This shows that both datasets are compatible and the merged dataset can further be used by the VLN community to evaluate their models on more diverse navigation instructions. Despite being trained on twice as many instances, the no image baseline still performs on par on map2seq unseen. From this we conclude that the current bottleneck for better generalization to unseen areas is the number of panorama images seen during training instead of number of instructions.

\section{Related Work}
Natural language instructed navigation of embodied agents has been studied in generated grid environments that allow a structured representation of the observed environment~\cite{walkthetalk, interpret_nlni}. Fueled by the advances in image representation learning~\cite{resnet}, the environments became more realistic by using real-world panorama images of indoor locations~\cite{room2room, rxr}. Complementary outdoor environments contain street level panoramas connected by a real-world street layout~\cite{city_without_map, Chen2018Touchdown, Mehta2020Retouchdown}. Agents in this outdoor environment are trained to follow human written navigation instructions~\cite{Chen2018Touchdown, Xiang2020LearningNavigation}, instructions generated by Google Maps~\cite{directions_in_streetview}, or a combination of both~\cite{Zhu2020MultimodalNavigation}. Recent work focuses on analyzing the navigation agents by introducing better trajectory overlap metrics~\cite{jain-etal-2019-stay, ndtw} or diagnosing the performance under certain constraints such as uni-modal inputs~\cite{Thomason2019ShiftingTB} and masking direction or object tokens~\cite{what_really_matters}. Other work used a trained VLN agent to evaluate automatically generated navigation instructions~\cite{zhao-etal-2021-evaluation}. An open problem in indoor VLN is the generalization of navigation performance to previously unseen areas. Proposed solutions include back translation with environment dropout~\cite{tan-etal-2019-learning}, multi-modal environment representation \cite{hu-etal-2019-looking} or semantic segmented images~\cite{Zhang2020DiagnosingNavigation}. Notably the latter work identifies the same problem in the Touchdown task.

\section{Conclusion}
We presented an investigation of outdoor vision and language navigation in seen and unseen environments. We introduced the heading delta feature and junction type embedding to correct an artifact of the environment and explicitly model the number of outgoing edges, respectively. Both are helpful to boost and analyze performance in the unseen scenario. We conducted experiments on two datasets and showed that the considered visual features poorly generalize to unseen areas. We conjecture that VLN tasks need to grow in scale and diversity of geographical environments and navigation tasks.

\subsubsection*{Acknowledgments}
The research reported in this paper was supported by a Google Focused Research Award on "Learning to Negotiate Answers in Multi-Pass Semantic Parsing".

\bibliography{anthology}
\bibliographystyle{acl_natbib}

\clearpage
\appendix

\section{Architecture Ablation}
We perform ablation studies on the ORAR full model in the seen scenario to measure the impact of individual architecture components. As seen in Table~\ref{tab:seen_ablation}, removing the second decoder RNN layer or BPE dropout results in a decrease of six and three task completion points, respectively. The largest drop in performance is observed when removing the text attention mechanism. This again shows the importance of attention over the encoder in sequence-to-sequence models. Removing the image attention mechanism on the other hand does not affect task completion on the map2seq dataset.
\begin{table}[]
\centering
\resizebox{.9\linewidth}{!}{
\begin{tabular}{lcclcc}
\toprule
     & \multicolumn{2}{c}{\textbf{Touchdown}} & \phantom{} & \multicolumn{2}{c}{\textbf{map2seq}} \\
\cmidrule{2-3}\cmidrule{5-6}
    \textbf{Ablation} & \textbf{dev}            & \textbf{test}          && \textbf{dev}          & \textbf{test}         \\
\midrule
ORAR full model         & \textbf{29.9}           & \textbf{29.1}          && \textbf{49.1}         & \textbf{46.7}         \\
\midrule
- no 2nd RNN             & 23.2          & 23.9          && 43.3         & 40.6         \\
- no BPE dropout         & 26.6          & 25.9          && 45.2         & 43.1         \\
- no text attention      & 9.4           & 10.4          && 22.0         & 21.8         \\
- no image attention     & 21.5          & 20.1          && \textbf{48.8}         & \textbf{45.7}\\
\bottomrule
\end{tabular}
}
\caption{ORAR full model ablation study on the \textbf{seen} scenario of Touchdown and map2seq. Metric is task completion and ablations are not cumulative.}
\label{tab:seen_ablation}
\end{table}

\section{Mixed-Model}
\begin{table*}[ht]
\centering
\resizebox{.99\linewidth}{!}{
\begin{tabular}{@{}lc@{}cc@{}clc@{}cc@{}clc@{}cc@{}clc@{}cc@{}c@{}}
\toprule
    &   \multicolumn{9}{c}{\textbf{Seen}} & \phantom{} &\multicolumn{9}{c}{\textbf{Unseen}}            \\ 
    
\cmidrule{2-10} \cmidrule{12-20}

    & \multicolumn{4}{c}{\textbf{Merged}} & \phantom{} & \multicolumn{2}{c}{\textbf{Touchdown}} & \multicolumn{2}{c}{\textbf{map2seq}} && 
       \multicolumn{4}{c}{\textbf{Merged}} & \phantom{} & \multicolumn{2}{c}{\textbf{Touchdown}} & \multicolumn{2}{c}{\textbf{map2seq}} \\ 
       
\cmidrule{2-5} \cmidrule{7-10} \cmidrule{12-15} \cmidrule{17-20}

    &  \multicolumn{2}{c}{\textbf{dev}} & \multicolumn{2}{c}{\textbf{test}} && \multicolumn{2}{c}{\textbf{test}} & \multicolumn{2}{c}{\textbf{test}} &&         \multicolumn{2}{c}{\textbf{dev}} & \multicolumn{2}{c}{\textbf{test}} && \multicolumn{2}{c}{\textbf{test}} & \multicolumn{2}{c}{\textbf{test}} \\ 
    
    \midrule
    
    \textbf{Model} & \multicolumn{1}{r}{\small{nDTW}} & TC & \multicolumn{1}{c}{\small{nDTW}} & TC && \multicolumn{1}{c}{\small{nDTW}} & TC & \multicolumn{1}{c}{\small{nDTW}} & TC && \multicolumn{1}{c}{\small{nDTW}} & TC & \multicolumn{1}{c}{\small{nDTW}} & TC && \multicolumn{1}{c}{\small{nDTW}} & TC & \multicolumn{1}{c}{\small{nDTW}} & TC \\
\midrule
best non-merged & - & - & - & - && 44.9 & 29.1 & 62.3 & 46.7 && - & - & - & - && 21.6 & 14.9 & 42.2 & 30.3 \\
best merged & 53.4 & 37.8 & 51.8 & 35.7 && 46.0 & 30.1 & \textbf{67.3} & \textbf{52.8} && \textbf{35.7} & \textbf{25.4} & \textbf{33.6} & \textbf{24.2} && \textbf{27.0} & \textbf{19.3} & \textbf{46.1} & \textbf{33.5} \\
\midrule
&\multicolumn{7}{c}{$\bullet$ 4th-to-last + pre-final} &&&& \multicolumn{7}{c}{$\bullet$ 4th-to-last + no image}&&\\
ORAR mixed model & \textbf{58.6} & \textbf{44.4} & \textbf{57.4} & \textbf{42.9} && \textbf{51.3} & \textbf{36.9} & - & - && \textbf{36.3} & \textbf{26.1} & \textbf{33.6} & \textbf{23.9} && \textbf{26.3} & \textbf{18.3} & - & - \\
\bottomrule
\end{tabular}
}
\caption{Results for the mixed model in comparison to previous best results. Metrics are normalized Dynamic Time Warping~(nDTW) and task completion~(TC). In the first two rows the best results of Table \ref{tab:results} and Table~\ref{tab:merged} are listed for comparison. The last section presents results for the \textit{ORAR mixed model} which uses different image features for different sub-tasks.}
\label{tab:mixed}
\end{table*}

The findings in Section~\ref{sec:oracle} inspire us to modify the ORAR model to use distinct visual features for the orientation and directions/stopping task. The orientation task is equivalent to the very first action prediction by the agent. Thus we modify the model architecture to use the ResNet 4th-to-last features (+text representation) to predict the first action and then start the recurrent prediction of the remaining actions with a different set of visual features~(pre-final for the seen scenario and no image features for the unseen scenario). The results for this \textbf{ORAR~mixed} model trained on the merged dataset are shown in Table~\ref{tab:mixed}. We only test it on Touchdown because map2seq does not have the orientation task. The mixed model significantly outperforms the single visual feature model on the Touchdown seen test set but unfortunately shows no improvement in the unseen scenario.

\section{Additional Metrics and Individual Runs}
We present the results of the individual repetitions and additional metrics for the main results in Table \ref{tab:results} and the results on the merged dataset in Table \ref{tab:merged}. The additional metrics are success weighted normalized Dynamic Time Warping \cite{ndtw} and shortest-path distance \cite{Chen2018Touchdown}.

\begin{table*}[]
\centering
\ra{1.05}
\resizebox{.99\linewidth}{!}{
\begin{tabular}{@{}lc@{}cc@{}clc@{}cc@{}clc@{}cc@{}clc@{}cc@{}c@{}}
\toprule
    &   \multicolumn{9}{c}{\textbf{Seen}} & \phantom{} &\multicolumn{9}{c}{\textbf{Unseen}}            \\ 
\cmidrule{2-10} \cmidrule{12-20}
    &   \multicolumn{4}{c}{\textbf{Touchdown}} & \phantom{} & \multicolumn{4}{c}{\textbf{map2seq}} && 
        \multicolumn{4}{c}{\textbf{Touchdown}} & \phantom{} & \multicolumn{4}{c}{\textbf{map2seq}} \\ 
\cmidrule{2-5} \cmidrule{7-10} \cmidrule{12-15} \cmidrule{17-20}
    &   \multicolumn{2}{c}{\textbf{dev}} & \multicolumn{2}{c}{\textbf{test}} && \multicolumn{2}{c}{\textbf{dev}} &         \multicolumn{2}{c}{\textbf{test}} && \multicolumn{2}{c}{\textbf{dev}} & \multicolumn{2}{c}{\textbf{test}} &&        \multicolumn{2}{c}{\textbf{dev}} & \multicolumn{2}{c}{\textbf{test}} \\ 
    \midrule
    \textbf{Model} & \multicolumn{1}{r}{\small{SDTW}} & SPD & \multicolumn{1}{c}{\small{SDTW}} & SPD && \multicolumn{1}{c}{\small{SDTW}} & SPD & \multicolumn{1}{c}{\small{SDTW}} & SPD && \multicolumn{1}{c}{\small{SDTW}} & SPD & \multicolumn{1}{c}{\small{SDTW}} & SPD && \multicolumn{1}{c}{\small{SDTW}} & SPD & \multicolumn{1}{c}{\small{SDTW}} & SPD \\
\midrule
RConcat         &  9.8 & 20.4 & 11.1 & 20.4 && 16.0 & 19.0 & 13.7 & 20.1 && 1.8 & 29.6 & 1.4 & 29.3 && 1.2 & 33.1 & 1.7 & 34.1 \\
GA              & 11.1 & 18.7 & 10.9 & 19.0 && 17.2 & 16.5 & 16.0 & 18.0 && 1.3 & 31.0 & 1.7 & 30.5 && 1.4 & 34.3 & 1.3 & 34.3 \\
ARC             & 14.1 & 18.6 & 13.5 & 19.4 && - & - & - & - && - & - & - & - && - & - & - & - \\
ARC+l2s         & 19.0 & 17.1 & 16.3 & 18.8 && - & - & - & - && - & - & - & - && - & - & - & - \\
VLN Transformer & 12.9 & 21.5 & 14.0 & 21.2 && 17.5 & 18.6 & 15.9 & 19.0 && 1.9 & 29.5 & 2.3 & 29.6 && - & - & - & - \\
\midrule
ORAR full model&&&&&&&&&&&&&&&&&&&\\
$\bullet$ ResNet pre-final & 24.5 & 15.0 & 23.8 & 16.2 && 46.7 & 5.9 & 44.4 & 6.6 && 8.6 & 26.7 & 7.6 & 26.7 && 22.3 & 15.6 & 22.8 & 16.3 \\
$\bullet$ ResNet 4th-to-last & 28.3 & 11.1 & 27.4 & 11.7 && 41.1 & 7.2 & 39.5 & 7.6 && 14.3 & 20.0 & 13.6 & 20.7 && 25.8 & 11.9 & 28.3 & 12.7 \\
\midrule
ORAR full model&\multicolumn{4}{c}{$\bullet$ ResNet 4th-to-last} & \phantom{} & \multicolumn{4}{c}{$\bullet$ ResNet pre-final} & \phantom{} & \multicolumn{4}{c}{$\bullet$ ResNet 4th-to-last} & \phantom{} & \multicolumn{4}{c}{$\bullet$ ResNet 4th-to-last} \\
- no heading delta & 28.3 & 10.9 & 27.6 & 11.5 && 45.4 & 6.8 & 42.7 & 7.7 && 14.0 & 20.5 & 13.5 & 20.8 && 20.4 & 16.8 & 21.9 & 17.1 \\
- no junction type & 23.1 & 13.6 & 22.8 & 13.9 && 47.2 & 7.6 & 43.0 & 8.6 && 4.0 & 26.6 & 3.7 & 26.7 && 4.3 & 28.9 & 4.2 & 29.9 \\
\bottomrule
\end{tabular}
}
\caption{Results on Touchdown and map2seq for the seen and unseen scenario. Metrics are success weighted normalized Dynamic Time Warping~(SDTW) and shortest-path distance~(SPD). For SDTW higher values are better and for SPD lower values are better.}
\label{tab:results_sdtw_sed}
\end{table*}

\begin{table*}[]
\centering
\ra{1.05}
\resizebox{.99\linewidth}{!}{
\begin{tabular}{@{}lrrrrrrrrrrrrlrrrrrrrrrrrr@{}}
\toprule
    &   \multicolumn{12}{c}{\textbf{Seen}} & \phantom{} &\multicolumn{12}{c}{\textbf{Unseen}}            \\ 
\cmidrule{2-13} \cmidrule{15-26}\\
&\multicolumn{10}{c}{\textbf{task completion of the ten repetitions}}&\textbf{mean}&\textbf{std}&\phantom{}&\multicolumn{10}{c}{\textbf{task completion of the ten repetitions}}&\textbf{mean}&\textbf{std}\\
\midrule
ORAR full model &&&&&&&&&&&&&&&&&&&&&&&&& \\
$\bullet$ ResNet pre-final & 26.1 & 18.5 & 25.8 & 25.1 & 26.8 & 28.7 & 24.4 & 25.5 & 25.6 & 26.0 & 25.3 & 2.5 && 8.8 & 9.2 & 7.3 & 9.8 & 8.5 & 8.4 & 10.0 & 8.2 & 9.4 & 8.1 & 8.8 & 0.8 \\
$\bullet$ ResNet 4th-to-last & 28.2 & 30.0 & 26.9 & 29.6 & 27.4 & 29.2 & 30.4 & 30.0 & 28.3 & 30.7 & 29.1 & 1.2 && 12.0 & 15.1 & 14.5 & 15.5 & 14.3 & 16.0 & 16.5 & 14.9 & 14.5 & 15.3 & 14.9 & 1.2 \\
\midrule
ORAR full model&\multicolumn{12}{c}{$\bullet$ ResNet 4th-to-last} & \phantom{} & \multicolumn{12}{c}{$\bullet$ ResNet 4th-to-last} \\
- no heading delta & 29.2 & 30.0 & 27.4 & 29.9 & 29.0 & 29.5 & 31.2 & 29.3 & 28.4 & 29.0 & 29.3 & 1.0 &&  14.7 & 13.7 & 15.5 & 14.9 & 14.1 & 13.5 & 16.0 & 15.1 & 16.0 & 14.5 & 14.8 & 0.8\\
- no junction type & 24.1 & 24.5 & 22.6 & 21.9 & 24.4 & 25.7 & 26.1 & 24.5 & 24.5 & 24.1 & 24.2 & 1.2 && 4.4 & 5.0 & 4.2 & 4.2 & 4.2 & 3.8 & 4.5 & 4.2 & 5.1 & 4.3 & 4.4 & 0.4 \\
\bottomrule
\end{tabular}
}
\caption{Task completion for the ten individual runs with mean and standard deviation on the Touchdown seen and unseen test set.}
\label{tab:results_10runs_td}
\end{table*}

\begin{table*}[]
\centering
\ra{1.05}
\resizebox{.99\linewidth}{!}{
\begin{tabular}{@{}lrrrrrrrrrrrrlrrrrrrrrrrrr@{}}
\toprule
    &   \multicolumn{12}{c}{\textbf{Seen}} & \phantom{} &\multicolumn{12}{c}{\textbf{Unseen}}            \\ 
\cmidrule{2-13} \cmidrule{15-26}\\
&\multicolumn{10}{c}{\textbf{task completion of the ten repetitions}}&\textbf{mean}&\textbf{std}&\phantom{}&\multicolumn{10}{c}{\textbf{task completion of the ten repetitions}}&\textbf{mean}&\textbf{std}\\
\midrule
ORAR full model &&&&&&&&&&&&&&&&&&&&&&&&& \\
$\bullet$ ResNet pre-final & 41.0 & 48.8 & 47.8 & 47.9 & 45.8 & 49.5 & 45.8 & 48.2 & 44.6 & 47.4 & 46.7 & 2.4 && 22.4 & 18.8 & 26.0 & 24.5 & 26.1 & 28.1 & 22.1 & 26.8 & 24.4 & 26.6 & 24.6 & 2.6 \\
$\bullet$ ResNet 4th-to-last & 40.5 & 42.2 & 42.1 & 42.1 & 38.6 & 42.9 & 41.2 & 42.1 & 45.2 & 40.5 & 41.7 & 1.6 && 32.9 & 29.6 & 28.9 & 28.5 & 27.6 & 32.2 & 26.8 & 33.6 & 34.0 & 28.4 & 30.3 & 2.5 \\
\midrule
ORAR full model&\multicolumn{12}{c}{$\bullet$ ResNet pre-final} & \phantom{} & \multicolumn{12}{c}{$\bullet$ ResNet 4th-to-last} \\
- no heading delta & 46.0 & 43.1 & 47.1 & 47.5 & 45.0 & 48.4 & 36.2 & 44.6 & 47.1 & 43.8 & 44.9 & 3.3 && 23.2 & 24.0 & 21.6 & 25.8 & 24.5 & 23.6 & 23.8 & 23.2 & 22.0 & 24.5 & 23.6 & 1.2\\
- no junction type & 44.9 & 46.1 & 46.2 & 44.0 & 43.2 & 46.5 & 44.9 & 47.1 & 45.5 & 42.1 & 45.1 & 1.5 && 5.1 & 4.5 & 5.0 & 5.1 & 4.6 & 3.9 & 5.6 & 3.8 & 4.4 & 4.6 & 4.7 & 0.5 \\
\bottomrule
\end{tabular}
}
\caption{Task completion for the ten individual runs with mean and standard deviation on the map2seq seen and unseen test set.}
\label{tab:results_10runs_m2s}
\end{table*}

\clearpage

\begin{table*}[]
\centering
\ra{1.05}
\resizebox{.99\linewidth}{!}{
\begin{tabular}{@{}lc@{}cc@{}clc@{}cc@{}clc@{}cc@{}clc@{}cc@{}c@{}}
\toprule
    &   \multicolumn{9}{c}{\textbf{Seen}} & \phantom{} &\multicolumn{9}{c}{\textbf{Unseen}}            \\ 
    
\cmidrule{2-10} \cmidrule{12-20}

    & \multicolumn{4}{c}{\textbf{Merged}} & \phantom{} & \multicolumn{2}{c}{\textbf{Touchdown}} & \multicolumn{2}{c}{\textbf{map2seq}} && 
       \multicolumn{4}{c}{\textbf{Merged}} & \phantom{} & \multicolumn{2}{c}{\textbf{Touchdown}} & \multicolumn{2}{c}{\textbf{map2seq}} \\ 
       
\cmidrule{2-5} \cmidrule{7-10} \cmidrule{12-15} \cmidrule{17-20}

    &  \multicolumn{2}{c}{\textbf{dev}} & \multicolumn{2}{c}{\textbf{test}} && \multicolumn{2}{c}{\textbf{test}} & \multicolumn{2}{c}{\textbf{test}} &&         \multicolumn{2}{c}{\textbf{dev}} & \multicolumn{2}{c}{\textbf{test}} && \multicolumn{2}{c}{\textbf{test}} & \multicolumn{2}{c}{\textbf{test}} \\ 
    
    \midrule
    
    \textbf{Model} & \multicolumn{1}{r}{\small{SDTW}} & SPD & \multicolumn{1}{c}{\small{SDTW}} & SPD && \multicolumn{1}{c}{\small{SDTW}} & SPD & \multicolumn{1}{c}{\small{SDTW}} & SPD && \multicolumn{1}{c}{\small{SDTW}} & SPD & \multicolumn{1}{c}{\small{SDTW}} & SPD && \multicolumn{1}{c}{\small{SDTW}} & SPD & \multicolumn{1}{c}{\small{SDTW}} & SPD \\
\midrule
ORAR full model\\
$\bullet$ no image          & 25.0 & 18.8 & 23.2 & 19.4 && 13.9 & 26.1 & 39.8 & 7.8 && 20.6 & 17.9 & 17.7 & 21.3 && 10.5 & 26.7 & 31.1 & 11.4 \\
$\bullet$ ResNet pre-final  & 36.8 & 12.5 & 34.8 & 14.1 && 26.1 & 18.8 & 50.2 & 5.7 && 20.3 & 20.1 & 18.4 & 22.0 && 12.2 & 25.8 & 30.2 & 14.8 \\
$\bullet$ ResNet 4th-to-last  & 35.9 & 9.3 & 33.8 & 9.8 && 28.4 & 11.7 & 43.2 & 6.5 && 23.6 & 14.9 & 22.5 & 16.6 && 17.7 & 19.2 & 31.4 & 11.7 \\
\midrule
ORAR mixed model\\
$\bullet$ 4th-to-last + pre-final  & 42.1 & 8.6 & 40.8 & 9.3 && 34.8 & 11.5 & - & - && - & - & - & - && - & - & - & - \\
$\bullet$ 4th-to-last + no image   & - & - & - & - && - & - & - & - && 24.1 & 15.1 & 22.2 & 17.2 && 16.9 & 20.4 & - & - \\
\bottomrule
\end{tabular}
}
\caption{Results for models trained on the merged dataset. Test results are presented for the merged test set and individual Touchdown and map2seq test sets. Metrics are success weighted normalized Dynamic Time Warping~(SDTW) and shortest-path distance~(SPD). For SDTW higher values are better and for SPD lower values are better.}
\label{tab:merged_sdtw_sed}
\end{table*}

\begin{table*}[]
\centering
\ra{1.05}
\resizebox{.99\linewidth}{!}{
\begin{tabular}{@{}lrrrrrrrrrrrrlrrrrrrrrrrrr@{}}
\toprule
    &   \multicolumn{12}{c}{\textbf{Seen}} & \phantom{} &\multicolumn{12}{c}{\textbf{Unseen}}            \\ 
\cmidrule{2-13} \cmidrule{15-26}\\
&\multicolumn{10}{c}{\textbf{task completion of the ten repetitions}}&\textbf{mean}&\textbf{std}&\phantom{}&\multicolumn{10}{c}{\textbf{task completion of the ten repetitions}}&\textbf{mean}&\textbf{std}\\
\midrule
ORAR full model &&&&&&&&&&&&&&&&&&&&&&&&& \\
$\bullet$ no image & 24.0 & 24.1 & 25.3 & 25.8 & 24.6 & 26.1 & 24.4 & 24.1 & 24.1 & 24.5 & 24.7 & 0.7 && 20.1 & 18.2 & 19.5 & 19.7 & 18.6 & 18.6 & 18.7 & 19.8 & 18.6 & 19.9 & 19.2 & 0.7\\
$\bullet$ ResNet pre-final & 36.7 & 34.7 & 35.3 & 37.1 & 36.4 & 36.1 & 38.8 & 36.4 & 36.8 & 39.5 & 36.8 & 1.4 && 18.9 & 19.8 & 19.8 & 20.8 & 20.1 & 20.0 & 20.5 & 19.9 & 20.2 & 19.9 & 20.0 & 0.5\\
$\bullet$ ResNet 4th-to-last & 34.9 & 36.2 & 35.4 & 35.4 & 36.2 & 36.5 & 34.1 & 36.3 & 36.3 & 35.5 & 35.7 & 0.7  && 23.8 & 24.9 & 25.8 & 23.9 & 24.1 & 23.4 & 24.7 & 23.9 & 23.5 & 24.2 & 24.2 & 0.7\\
\midrule
ORAR mixed model &&&&&&&&&&&&&&&&&&&&&&&&& \\
$\bullet$ 4th-to-last + pre-final & 43.8 & 42.6 & 43.4 & 43.2 & 43.6 & 42.0 & 44.1 & 42.1 & 41.9 & 42.7 & 42.9 & 0.8 && - & - & - & - & - & - & - & - & - & - & - & -\\
$\bullet$ 4th-to-last + no image & - & - & - & - & - & - & - & - & - & - & - & - && 23.8 & 23.4 & 23.7 & 23.4 & 24.1 & 24.7 & 24.6 & 24.1 & 23.8 & 22.9 & 23.8 & 0.5\\
\bottomrule
\end{tabular}
}
\caption{Task completion for the ten individual runs with mean and standard deviation on the merged seen and unseen test set.}
\label{tab:merged_10runs_merged}
\end{table*}

\begin{table*}[]
\centering
\ra{1.05}
\resizebox{.99\linewidth}{!}{
\begin{tabular}{@{}lrrrrrrrrrrrrlrrrrrrrrrrrr@{}}
\toprule
    &   \multicolumn{12}{c}{\textbf{Seen}} & \phantom{} &\multicolumn{12}{c}{\textbf{Unseen}}            \\ 
\cmidrule{2-13} \cmidrule{15-26}\\
&\multicolumn{10}{c}{\textbf{task completion of the ten repetitions}}&\textbf{mean}&\textbf{std}&\phantom{}&\multicolumn{10}{c}{\textbf{task completion of the ten repetitions}}&\textbf{mean}&\textbf{std}\\
\midrule
ORAR full model &&&&&&&&&&&&&&&&&&&&&&&&& \\
$\bullet$ no image & 14.1 & 14.4 & 15.8 & 16.5 & 14.1 & 16.3 & 14.5 & 14.9 & 13.3 & 14.5 & 14.8 & 1.0 && 12.1 & 10.7 & 12.1 & 12.2 & 11.0 & 11.5 & 11.5 & 12.9 & 11.0 & 12.2 & 11.7 & 0.7\\
$\bullet$ ResNet pre-final & 27.5 & 25.1 & 26.4 & 28.4 & 26.6 & 27.4 & 30.3 & 27.7 & 27.0 & 30.2 & 27.7 & 1.5 && 13.1 & 13.0 & 13.1 & 14.1 & 13.5 & 13.7 & 14.1 & 13.5 & 13.9 & 13.7 & 13.6 & 0.4\\
$\bullet$ ResNet 4th-to-last & 30.7 & 30.4 & 30.0 & 30.1 & 30.0 & 30.2 & 29.2 & 30.2 & 30.4 & 30.0 & 30.1 & 0.4 && 18.0 & 20.3 & 20.8 & 18.8 & 18.9 & 18.0 & 20.2 & 19.8 & 18.6 & 19.6 & 19.3 & 0.9\\
\midrule
ORAR mixed model &&&&&&&&&&&&&&&&&&&&&&&&& \\
$\bullet$ 4th-to-last + pre-final & 37.6 & 36.3 & 36.4 & 37.8 & 37.9 & 35.1 & 38.0 & 36.6 & 35.6 & 37.4 & 36.9 & 1.0 && - & - & - & - & - & - & - & - & - & - & - & -\\
$\bullet$ 4th-to-last + no image & - & - & - & - & - & - & - & - & - & - & - & - && 17.8 & 17.9 & 18.1 & 18.8 & 18.8 & 19.2 & 19.2 & 18.4 & 17.9 & 17.1 & 18.3 & 0.6\\
\bottomrule
\end{tabular}
}
\caption{Task completion for the ten individual runs with mean and standard deviation on the Touchdown seen and unseen test set, trained on the merged training set.}
\label{tab:merged_10runs_td}
\end{table*}

\begin{table*}[]
\centering
\ra{1.05}
\resizebox{.99\linewidth}{!}{
\begin{tabular}{@{}lrrrrrrrrrrrrlrrrrrrrrrrrr@{}}
\toprule
    &   \multicolumn{12}{c}{\textbf{Seen}} & \phantom{} &\multicolumn{12}{c}{\textbf{Unseen}}            \\ 
\cmidrule{2-13} \cmidrule{15-26}\\
&\multicolumn{10}{c}{\textbf{task completion of the ten repetitions}}&\textbf{mean}&\textbf{std}&\phantom{}&\multicolumn{10}{c}{\textbf{task completion of the ten repetitions}}&\textbf{mean}&\textbf{std}\\
\midrule
ORAR full model &&&&&&&&&&&&&&&&&&&&&&&&& \\
$\bullet$ no image & 41.5 & 41.2 & 42.1 & 42.1 & 43.2 & 43.4 & 41.9 & 40.4 & 43.1 & 42.1 & 42.1 & 0.9 && 35.1 & 32.5 & 33.6 & 33.8 & 32.9 & 32.0 & 32.4 & 32.8 & 32.8 & 34.2 & 33.2 & 0.9\\
$\bullet$ ResNet pre-final & 53.0 & 51.5 & 51.0 & 52.4 & 53.6 & 51.5 & 53.6 & 51.6 & 53.9 & 55.8 & 52.8 & 1.4 && 29.9 & 32.6 & 32.2 & 33.4 & 32.6 & 32.0 & 32.6 & 32.0 & 32.0 & 31.5 & 32.1 & 0.9\\
$\bullet$ ResNet 4th-to-last & 42.5 & 46.4 & 44.9 & 44.6 & 47.1 & 47.8 & 42.8 & 47.1 & 46.6 & 45.2 & 45.5 & 1.7 && 34.8 & 33.5 & 35.2 & 33.5 & 33.9 & 33.6 & 33.1 & 31.6 & 32.6 & 33.0 & 33.5 & 1.0\\
\bottomrule
\end{tabular}
}
\caption{Task completion for the ten individual runs with mean and standard deviation on the map2seq seen and unseen test set, trained on the merged training set.}
\label{tab:merged_10runs_m2s}
\end{table*}

\end{document}